\documentclass[authoryear,preprint,review,12pt]{elsarticle}

\usepackage{lineno,hyperref}
\usepackage{amsmath}
\usepackage{xcolor}
\usepackage{subcaption}

\journal{Neural Networks}

\begin{document}

\begin{frontmatter}

\title{
%Improving the performance of deep networks with non-negative matrix factorization \\
%Local inhibition improves the performance of deep networks composed of non-negative matrix factorization modules\\ \\
%Local mixing improves performance of deep non-negative matrix factorization networks. \\
%***
%\\
%Non-negative matrix factorization with local mixing improves performance of deep networks. \\
%***
%\\
Including local feature interactions in deep non-negative matrix factorization networks improves performance. 
%Hyper-columns in deep convolutional networks improve performances. \\
%***
%\\
%Improving deep networks with hyper-columns. 
}

\author[label1]{Mahbod Nouri\corref{mycorrespondingauthor}}
\ead{mahbod@uni-bremen.de}
\author[label1]{David Rotermund}
\author[label2]{Alberto Garcia-Ortiz}
\author[label1]{Klaus R. Pawelzik}
\cortext[mycorrespondingauthor]{Corresponding author}

\affiliation[label1]{organization={University of Bremen},
             addressline={Institute for Theoretical Physics},
             city={Bremen},
             postcode={28359},
             state={Bremen},
             country={Germany}}

 \affiliation[label2]{organization={University of Bremen},
             addressline={Institute of Electrodynamics and Microelectronics (ITEM.ids)},
             city={Bremen},
             postcode={28359},
             state={Bremen},
             country={Germany}}

\begin{abstract}

The brain uses positive signals as a means of signaling. Forward interactions in the early visual cortex are also positive, realized by excitatory synapses. Only local interactions also include inhibition. Non-negative matrix factorization (NMF) captures the biological constraint of positive long-range interactions and can be implemented with stochastic spikes. While NMF can serve as an abstract formalization of early neural processing in the visual system, the performance of deep convolutional networks with NMF modules does not match that of CNNs of similar size. However, when the local NMF modules are each followed by a module that mixes the NMF's positive activities, the performances on the benchmark data exceed that of vanilla deep convolutional networks of similar size. This setting can be considered a biologically more plausible emulation of the processing in cortical (hyper-)columns with the potential to improve the performance of deep networks. 

% Beside higher performance in networks of moderate size this novel approach promises also larger noise robustness and what else ...?

\end{abstract}

\begin{keyword}
deep neuronal networks \sep non-negative matrix factorization (NMF) \sep backpropagation error learning 
\end{keyword}

\end{frontmatter}

\section{Introduction}

% Already this insight made deep networks of NNMFs much more powerful, as we show below. 

% We then asked ourselves if more generally deep networks might profit from including NNMFs

% Only after all these motivating words the NNMFs should be explained in the methods section...

The success of modern neural networks has often come from incorporating principles inspired by their biological role model: the brain. A well-known example is the introduction of convolutional layers \citep{CNN}, which mirror the organization of the visual cortex in biological brains. Neurons in the visual cortex have localized receptive fields, meaning they respond stereotypically to stimuli in specific regions of the visual field \citep{Hubel1962}. This principle first adopted by \citep{Fukushima1980}, now enables convolutional neural networks (CNNs) to efficiently detect patterns and hierarchies in images. This biologically inspired design has been crucial in advancing the performance of machine learning models in computer vision \citep{726791}.

Despite the success of incorporating some biological principles, several key constraints in biological neural systems remain under-explored in machine learning. For example, Dale’s Law, one of the core principles of neurobiology, dictates that neurons are either excitatory or inhibitory but not both \citep{dale}. 

Another often overlooked biological aspect is the nature of long-range connections between cortical areas, such as those between the primary visual cortex (V1) and higher visual areas. In biological neural systems, long-range connections are predominantly excitatory, in contradiction with usual deep CNNs. In the cortex, the excitatory postsynaptic currents (EPSCs) are also received by inhibitory neurons, which in turn locally modulate pyramidal (Pyr) cells through inhibitory postsynaptic currents (IPSCs) \citep{Yang2013}, typically in a different layer. This configuration implies that while interactions of both signs are essential for modulating neural responses and shaping information processing locally, the primary flow of information across layers and areas is governed by excitatory or "positive" connections. 

% Here, we furthermore consider also the hypothesis that the receptive fields of neurons in the visual cortex emerge through sparse coding \citep{olshausen2006other} — a process where neural representations are formed by activating only a small subset of neurons. Non-negative matrix factorization (NMF) \citep{Lee1999} \citep{lee2000algorithms} captures both the biologically realistic constraints of the positivity of long-range interactions and the tendency to yield sparse representations. Importantly, it can be realized with stochastic spikes [CITE]. For instance, \citep{HOYER2003547} demonstrated that such sparse coding mechanisms could be effectively modeled using Non-negative Matrix Factorization (NMF). By imposing both non-negativity and sparseness constraints, this work showed that NMF can learn parts-based, interpretable representations similar to the mechanism yielding receptive fields first proposed in \citep{Olshausen1996}.   

Our work also builds on the hypothesis that receptive fields in the visual cortex develop through sparse coding mechanisms—where neural activity is distributed such that only a small subset of neurons responds to a given stimulus \citep{olshausen2006other}. Non-negative matrix factorization (NMF) \citep{Lee1999} \citep{lee2000algorithms} provides an elegant mathematical framework that simultaneously satisfies two key biological constraints: the positivity of long-range neural interactions and the tendency toward sparse representations. For instance, \citep{HOYER2003547} demonstrated that such sparse coding mechanisms could be effectively modeled using NMF. By imposing both non-negativity and sparseness constraints, this work showed that NMF can learn parts-based, interpretable representations similar to the mechanism yielding receptive fields first proposed in \citep{Olshausen1996}.

% This approach can be implemented using stochastic spike-based mechanisms [CITE]. 

While Non-negative Matrix Factorization (NMF) is a powerful unsupervised learning technique used in various fields, including signal processing, computer vision, and data mining \citep{Lee1999}, applying it effectively to supervised tasks like computer vision presents significant challenges. When used in isolation, NMF typically cannot match the performance of modern deep learning architectures such as CNNs. Previous works have attempted to bridge this gap by combining NMF with deep learning approaches. 

For instance, \citep{hamburger} proposed using an NMF layer on top of a convolutional model. However, such implementations often reinitialize and retrain the NMF components from scratch after each forward pass, leading to computational inefficiency and potential instability. In contrast, our approach implements a hierarchical NMF architecture where the NMF weights are treated as learnable parameters and optimized through back-propagation alongside the network's other parameters. This enables the NMF components to adapt continuously to the task requirements while maintaining their biological constraints.

Missing in current networks using NMF are local interactions that include inhibition, an important property of cortical microcircuits. 
After briefly reviewing NMF, we introduce a convolutional network architecture where we exchange CNN modules with NMF modules. We then propose a simple but novel extension of the NMF network where subsequent 1x1 convolutional layers are inserted. Thereby, we realize general local interactions among the features of the NMF modules in analogy to cortical hyper-columns, which makes these networks a step toward more biologically realistic models. When optimized with back-propagation (both the 1x1 CNNs and the NMF modules), we show that these networks exhibit performances on benchmark data sets that can exceed the values of pure CNNs with the same architecture. 

%NMF decomposes a non-negative matrix $V$ into two non-negative matrices $W$ and $H$, such that $V \approx WH$. This decomposition aims to find a parts-based representation of the data, where each component corresponds to a meaningful feature or pattern \citep{lee2001algorithms}. 

%NMF offers several key advantages in data analysis and representation. It provides faithful and interpretable representations of inherently non-negative data, ensuring accuracy and ease of understanding. Its tendency towards sparse representations contributes to robustness against noise, enhancing reliability in data processing. Additionally, the positivity constraint in NMF makes it particularly well-suited for modeling systems with naturally positive activities and weights, such as neural networks with spiking neurons. Given these properties, NMF is often used for data analysis and dimensionality reduction and has applications in fields such as image processing, text mining, and bio-informatics \citep{Buciu}. 

%\underline {Link between NMF and spiking models?}

\section{Methods}
\subsection{Non-negative Matrix Factorization (NMF)}
Non-negative Matrix Factorization (NMF) \citep{Lee1999} \citep{lee2000algorithms}  is a technique used to decompose a non-negative matrix $X$ of $M$ input vectors into two lower-dimensional non-negative matrices $W$ and $H$, such that $X \approx W H$, where $X \in R^{M \times S} \geq 0$,  $W \in R^{S \times I} \geq 0$, and $H \in R^{M \times I} \geq 0$. The goal is to minimize the difference between $X$ and the product $W H$ while ensuring that both $W$ and $H$ remain non-negative. It is often expressed as:

\begin{equation}
\min_{W, H} \| X - W H \| \quad \text{subject to} \quad (W)_{s j} = W_{s j} \geq 0, \, (H)_{\mu j} =: h^\mu_j\geq 0
\end{equation}

% Commonly, Euclidean distance 

% \begin{equation}
% \| X - WH \| = \sqrt{\sum_{i,j} (X_{ij} - (WH)_{ij})^2}
% \end{equation}

minimizing the Kullback-Leibler divergence defined as: 

\begin{equation}
D(X||W H) = \sum_{\mu, s} X_{\mu s} \ln{\frac{X_{\mu s}}{\sum_j W_{sj} h^\mu_j}}
\end{equation}

leads to the following multiplicative update rules for $W$ and $H$ \citep{lee2000algorithms}:

\begin{equation}
\label{update_h}
h^\mu_i \leftarrow h^\mu_i \sum_{s} \frac{W_{si} X_{\mu s}}{\sum_j W_{sj}h^\mu_j}
\end{equation}

\begin{equation}
W_{si} \leftarrow W_{si} \sum_{\mu} \frac{h^\mu_i X_{\mu j}}{\sum_j W_{sj}h^\mu_j}
\end{equation}

\begin{equation}
    W_{si} \leftarrow \frac{W_{si}}{\sum_{j} W_{ji}}
\end{equation}

% \underline{Normalization?}
\subsection{Deep Non-negative Matrix Factorization in a Neural Network}

In this work, we extend NMF to a deep learning setting \citep{Chen2022} by integrating it into a network architecture. Specifically, we treat the factorized matrices $W$ and $H$ as components of a neural network layer. The matrix $W$ is used as the weight matrix of the layer, while the matrix $H$ represents the activation values (neuron outputs) of the layer. 

The challenge lies in adapting the unsupervised nature of NMF \citep{Lee1999} for use in a multi-layer supervised context, such as classification, where the learned weights must optimize a specific task-related objective function \citep{ciampiconi2023survey} \citep{tian2022recent}. In classical NMF, both $W$ and $H$ are updated iteratively using multiplicative update rules, following the Expectation-Maximization algorithm, to minimize the factorization error. However, when applying NMF within a neural network, directly updating $W$ in an unsupervised manner could lead to weights that are not aligned with the task objective (e.g., classification loss). To address this, we decouple the update process for $W$ from the factorization step.

Instead of updating $W$ using the NMF update rules, we keep $W$ fixed during the forward pass, using it to calculate the activations $H$ for each layer. The neuron values at each iteration, on the other hand, follow a similar approach to the NMF update rule. For one pattern the general update rule for $h$ in a hidden layer at each iteration $t$ is formulated as:
\begin{equation}
        h_i(t) = h_i(t-1) + \varepsilon h_i(t-1) \left(\sum_s\frac{X_s W_{s,i}}{ \sum_i W_{s,i}h_i(t-1)} - 1\right) 
    \label{update_h_ours}
\end{equation}

Where $X_s$ denotes the input and $W_{s,i}$ is the weight matrix. Unless said otherwise, $\varepsilon$ is set to 1, which leads to an equation similar to \ref{update_h}. During each forward pass at each layer, we first initialize $h$ values to $h_i(0) = \frac{1}{I}$, where $I$ is the number of neurons, and we repeat the update rule (\ref{update_h_ours}) for $N$ times until getting the output values of the layer.

\subsection{Approximated back-propagation}
\begin{figure}[htp]
    \centering
    \includegraphics[width=\textwidth]{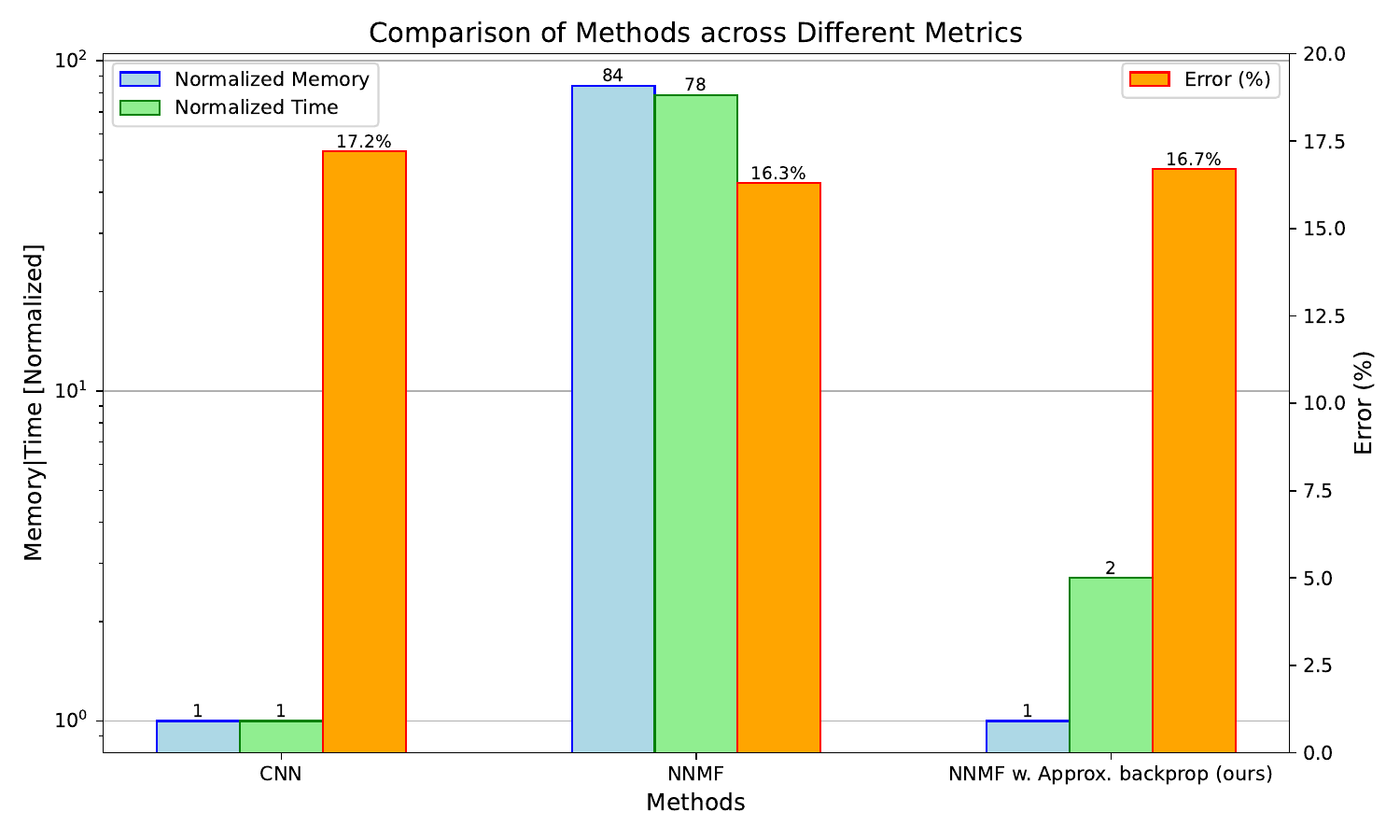}
    \caption{Performance and computational cost comparison during the back-propagation between CNN, NMF, and NMF with approximate back-propagation (ours). The comparison spans three metrics: back-propagation memory consumption (left), back-propagation computation time (middle), and classification error (right). Memory and time values are shown relative to the CNN baseline.}
    \label{approx_compare}
\end{figure}

\begin{figure}[htp]
    \centering
    \includegraphics[width=\textwidth]{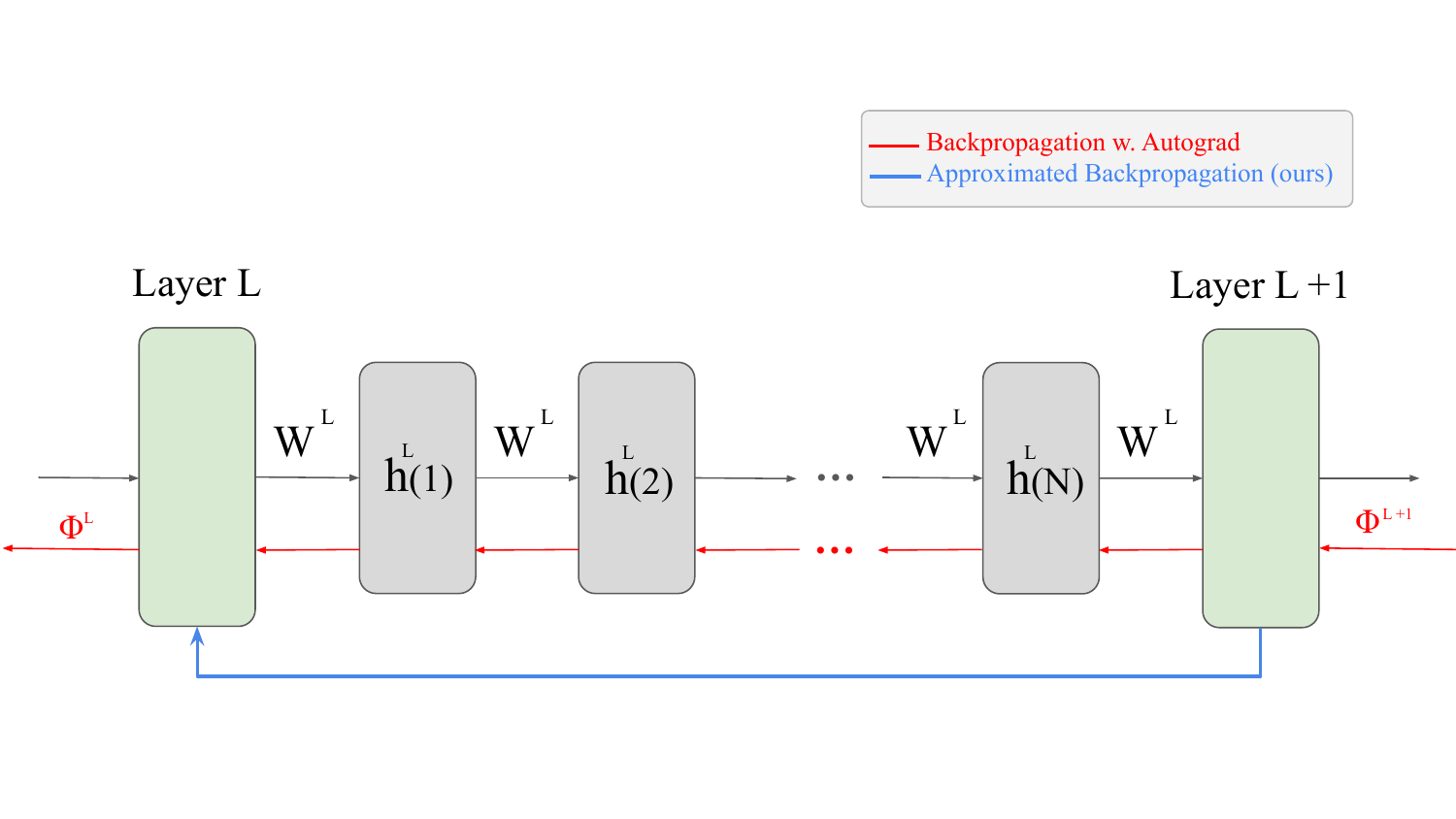}
    \caption{The difference between our approximative approach and the naive back-propagation. Since NMF is an iterative algorithm, the output of each layer is computed after several iterations of the update rule. To apply the vanilla back-propagation, all these intermediate steps are required to be saved to the memory during the forward pass, which is time- and memory-inefficient. Instead, our proposed approximated back-propagation can compute the corresponding error of a lower layer in one step, only utilizing the output of the layer.}
    \label{approx}
\end{figure}

A significant practical challenge in implementing NMF-based neural architectures is the computational overhead of back-propagation through iterative steps. Conventional NMF requires $N$ iterations ($N$ usually ranging from 20 to 100 iterations) in the forward pass, and automatic differentiation frameworks like PyTorch must store gradients for each iteration to compute the backward pass accurately. This creates substantial memory requirements and computational bottlenecks, especially for deep networks or large-scale applications. 

We address this limitation by using an efficient approximation to the back-propagation procedure that requires only a single step, eliminating the need to save and back-propagate through all intermediate iterations performed during the forward pass. This method introduced in \citep{rotermund2019back}  dramatically reduces both memory consumption and computation time while maintaining comparable accuracy to full back-propagation through all iterations. For convenience we here only sketch the basic idea underlying this approach but refer to \citep{rotermund2019back} for the detailed derivations. 
%%% KLAUS EQUATIONS:
Back-propagation requires the partial derivatives $\frac{\partial h_i}{\partial x_s}$ and $\frac{\partial h_i}{\partial W_{s,j}}$. These derivatives can be obtained from the the h-dynamics for a single pattern $x$:  

\begin{eqnarray}
     h'_i &=& h_i + \varepsilon h_i \left(\sum_s\frac{x_s W_{s,i}}{R_s} - 1 \right) \\
    &=& h_i + \delta h_i,
 \end{eqnarray}
where $R_s := \sum_i W_{s,i}h_i$.
%Now let's assume this dynamics is in a fixed point $h(t)$ that is $\Delta h_i = 0$. 
If we would change $x \to x + \Delta x$ we would obtain a deterministic change of the output in one step of this dynamics: 

 \begin{eqnarray}
    h'_i &=& h_i 
     + \varepsilon h_i \left(\sum_s\frac{(x_s + \Delta x_s) W_{s,i}}{R_s} - 1\right) \\
     &=& h_i 
     + \varepsilon h_i \left(\sum_s\frac{x_s W_{s,i}}{R_s} 
     + \sum_s\frac{\Delta x_s W_{s,i}}{R_s} - 1 \right) \\
     %&=& 
     %h_i     + \varepsilon h_i\sum_s\frac{\Delta x_s W_{s,i}}{R_s} \\
     &=& h_i + \delta h_i + \Delta h_i.
 \end{eqnarray}
 That is, we now have the changes of the original $\delta h$ depending on changes of the input $\Delta x$:
 \begin{eqnarray}
     \Delta h_i &=& \varepsilon h_i \left(\sum_s\frac{\Delta x_s W_{s,i}}{R_s} \right).
 \end{eqnarray}
 This formula preserves normalization of $h$ since $\sum_i \Delta h_i = 0$. \\ \\
 By comparing with the total differential, we obtain 
 \begin{eqnarray}
     \frac{\partial h'_i}{\partial x_s} &\propto& h_i\frac{ W_{s,i}}{R_s} .
 \end{eqnarray}

% Based on this equation, we can write the back-propagation error for layer $L-1$ based on incoming error $\Phi$ and parameters in layer $L$ as:
% \begin{equation}
%     \Phi_{L-1}(s) = \frac{\sum_i^N \Phi_{L}(i) h(i) W(s|i)}{R_s} 
%     \label{eq:newbackproperror}
% \end{equation}

%\textbf{KLAUS EQUATIONS:}
Following the same logic we have
\begin{eqnarray}
    h_i' = h_i + \varepsilon (\sum_s\frac{x_s (W_{s,i} + \Delta W_{s,i})}{\sum_j (W_{s,j} + \Delta W_{s,j}) h_j} - 1) \\
    \simeq 
    h_i + \delta h_i + \varepsilon h_i (\sum_s\frac{x_s (W_{s,i} + \Delta W_{s,i})}{R_s} 
    - \sum_s\frac{x_s (W_{s,i} + \Delta W_{s,i}) (\sum_j \Delta W_{s,j} h_j)}{R_s^2} ) \\
    \approx
    h_i + \delta h_i + \varepsilon h_i (\sum_s\frac{x_s \Delta W_{s,i}}{R_s} 
    - \sum_s\frac{x_s (W_{s,i}) (\sum_j \Delta W_{s,j} h_j)}{R_s^2}).
\end{eqnarray}
which (again via the total differential) leads to 

\begin{equation}
    \frac{\partial h'_i}{\partial W_{s,j}} =
    h_i \frac{x_s}{R_s}(\delta_{i,j} - \frac{W_{s,i} h_j}{R_s}).
\end{equation}

%% DAVID'S EQUATIONS:

%\textbf{DAVID EQUATIONS:}
%The back-propagation learning algorithm is described in detail in \citep{Rotermund_19}. Here only a summary: 
%\begin{eqnarray}
%r_\mu(s,i) &=& W(s|i) h_\mu(i) \\
%R_\mu(s) &=& \sum_i r_\mu(s,i) \\
%z_\mu(s,i) &=&  \frac{r_\mu(s,i) }{R_\mu(s) + 10^{-20}} \\
%\Phi_\mu(s) &=& \sum_i \Phi_\mu^{\mbox{previous layer}}(i) z_\mu(s,i) \\
%f_\mu(s,i) &=& \frac{h_\mu(i) p_\mu(s)}{R_\mu^2(s) + 10^{-20} } \\
%\delta \omega(s|i) &=& \sum_\mu \left[R_\mu(s) h_\mu(i) - \sum_i r_\mu(s,i) \Phi_\mu^{\mbox{previous layer}}(i)\right] f_\mu(s,i)
%\end{eqnarray}

%\textbf{Or in more humanly-readable terms:}
These derivatives are then used to change the weights according to 
\begin{equation}
    \label{delta_u}
    \delta \omega_{si} = \frac{h_i X_s}{(R_s)^2} (\Phi_i^{L+1}R_s  - \sum_j W_{s j} h_j \Phi_j^{L+1})
\end{equation}
where 
\begin{equation}
    \Phi_s^L = \sum_i \Phi_i^{L+1} \frac{W_{si} h_i}{\sum_j W_{sj} h_j} \\
\end{equation}
is the back-propagated error $\Phi_i^{L+1}$. This method is applied  only to the final states $h(N)$. Figure \ref{approx} illustrates how this approach reduces the amount of computations.

\subsubsection{Updating the Weight Matrix}

To make the weight matrix $W$ more suitable for classification, we update $W$ through back-propagation using an optimizer (e.g., Adam). This ensures that $W$ is optimized based on the task's objective function rather than simply minimizing the reconstruction error from NMF.
However, a challenge arises as gradient-based updates do not inherently preserve the non-negativity and normalization properties of $W$. Since NMF requires that $W$ remains non-negative and normalized, we cannot directly update $W$ with the raw gradient values. Instead, we introduce a trainable auxiliary matrix $U$, which has the same dimensions as $W$, and at each network update, the optimizer will update $U$ using the error calculated in the equation \ref{delta_u}. Based on this parameter, during each forward pass, the weight $W_{s,i}$ is obtained based on:

\begin{equation}
    W_{s,i} = \frac{|U_{s,i}|}{\sum_k^S |U_{k,i}|}
\end{equation}

which applies two main transformations:
\begin{enumerate}
    \item \textbf{Non-negativity constraint}: We enforce non-negativity by setting $ W = |U| $, where $ |U| $ represents the element-wise absolute value of $ U $.
    \item \textbf{Normalization}: We normalize each row of $W$ to ensure the sum of each row is equal to 1, ensuring that $W$ remains a valid factorization matrix.
\end{enumerate}
 
The transformation ensures that the weight matrix $W$ retains the necessary properties for NMF while still being adaptable for learning tasks.

Our empirical evaluation confirms the computational efficiency of the proposed approximate back-propagation (BP) method while maintaining performance. Figure \ref{approx_compare} compares three architectures: a normal convolutional network, an NMF-based network with full BP via Torch Autograd, and our NMF network with approximate BP. While the standard NMF implementation shows considerable computational costs, requiring significantly more memory and time compared to the CNN baseline, our approximation method dramatically reduces these overheads. Specifically, while achieving comparable classification accuracy to both baseline models, our approximate BP approach maintains the same memory footprint as the CNN model while operating the BP $\approx 29$ times faster than the standard NMF. 

These results demonstrate that our approximation strategy successfully addresses the primary computational bottleneck of NMF-based networks while preserving their advantages. This computational innovation makes NMF-based neural architectures more practical for real-world applications, allowing us to leverage their biological plausibility advantages without prohibitive computational costs.

\subsection{Proposed Methods}
\begin{figure}[htp]
    \centering
    \includegraphics[width=\textwidth]{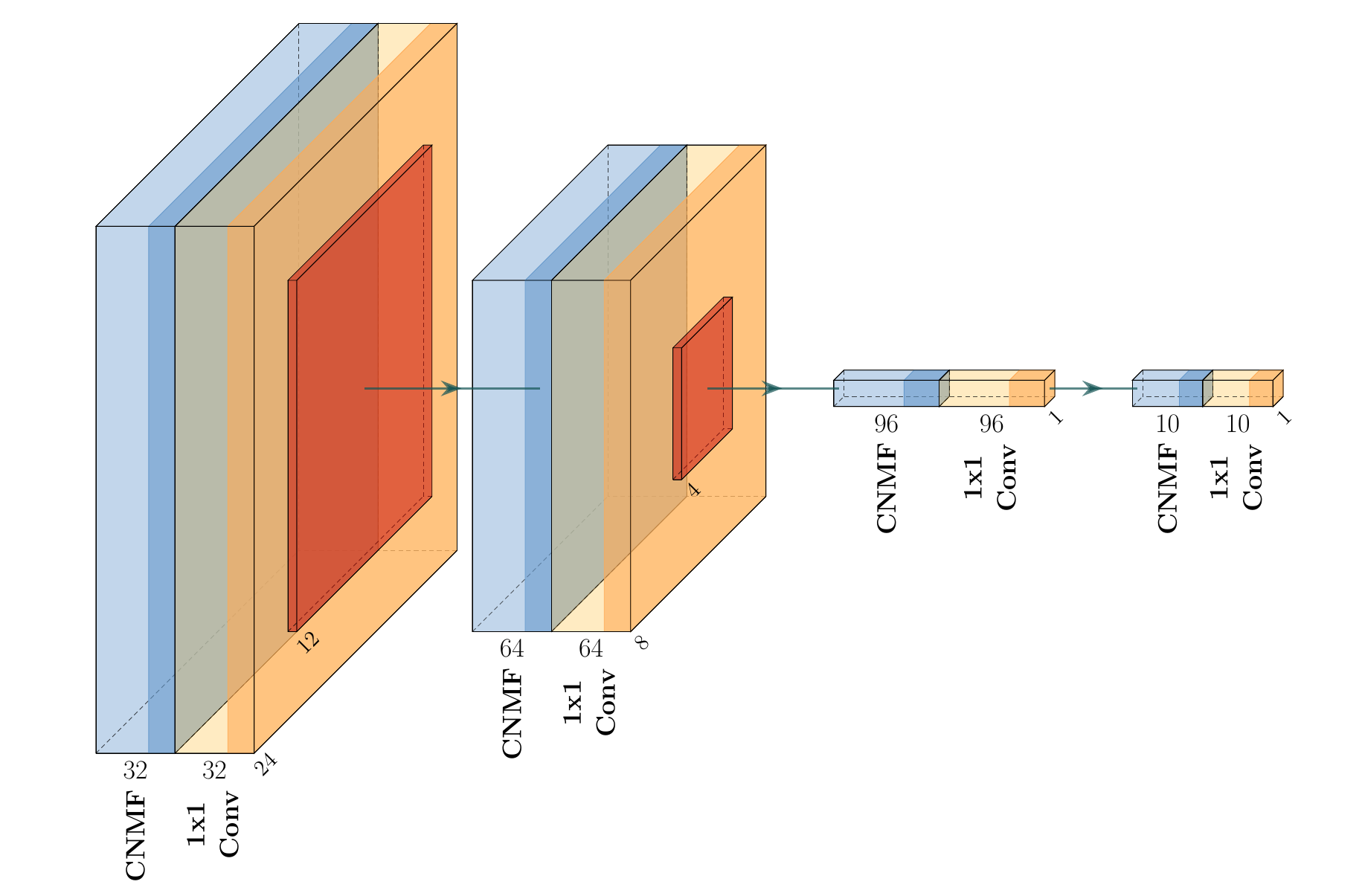}
    \caption{Network architecture of the proposed method for the \textbf{CNMF + $1\times 1$ Convolution}. The network consists of four sequential blocks, each containing a CNMF module followed by a 1$\times$1 convolutional layer. The architecture progressively reduces spatial dimensions from 28×28 in the input to 1×1 while transforming feature channels ($32\rightarrow64\rightarrow96\rightarrow10\rightarrow$ output). The output of the last 1$\times$1 convolutional layer is used for the classification. For simplicity, activations and batch normalization layers are omitted from the figure.}
    \label{model}
\end{figure}

\subsubsection{Convolutional NMF (CNMF)}
In conventional NMF \citep{lee2000algorithms}, the input is reconstructed using a linear transformation of the latent values, implemented through regular matrix multiplication, which corresponds to a dense layer in a neural network architecture. However, this linear transformation can be replaced with other linear operations while preserving the core principles of NMF.

In our approach, we substitute the standard matrix multiplication with a convolution operation, resulting in Convolutional NMF (CNMF). This adaptation maintains the mathematical foundations of NMF while leveraging the spatial locality benefits of convolutions. As demonstrated in our previous work \citep{Competitive}, CNMF can be effectively trained using back-propagation and exhibits remarkable noise robustness when compared to conventional CNNs.

While CNMF shows superior performance in noisy conditions, on clean data it does not consistently outperform standard CNNs with comparable architectures. To address this limitation and further enhance the capabilities of our CNMF approach, we propose an extended architecture incorporating additional components as described in the following sections.

\subsubsection{1$\times$1 Convolutions}
\begin{figure}[htp]
    \centering
    \includegraphics[width=\textwidth]{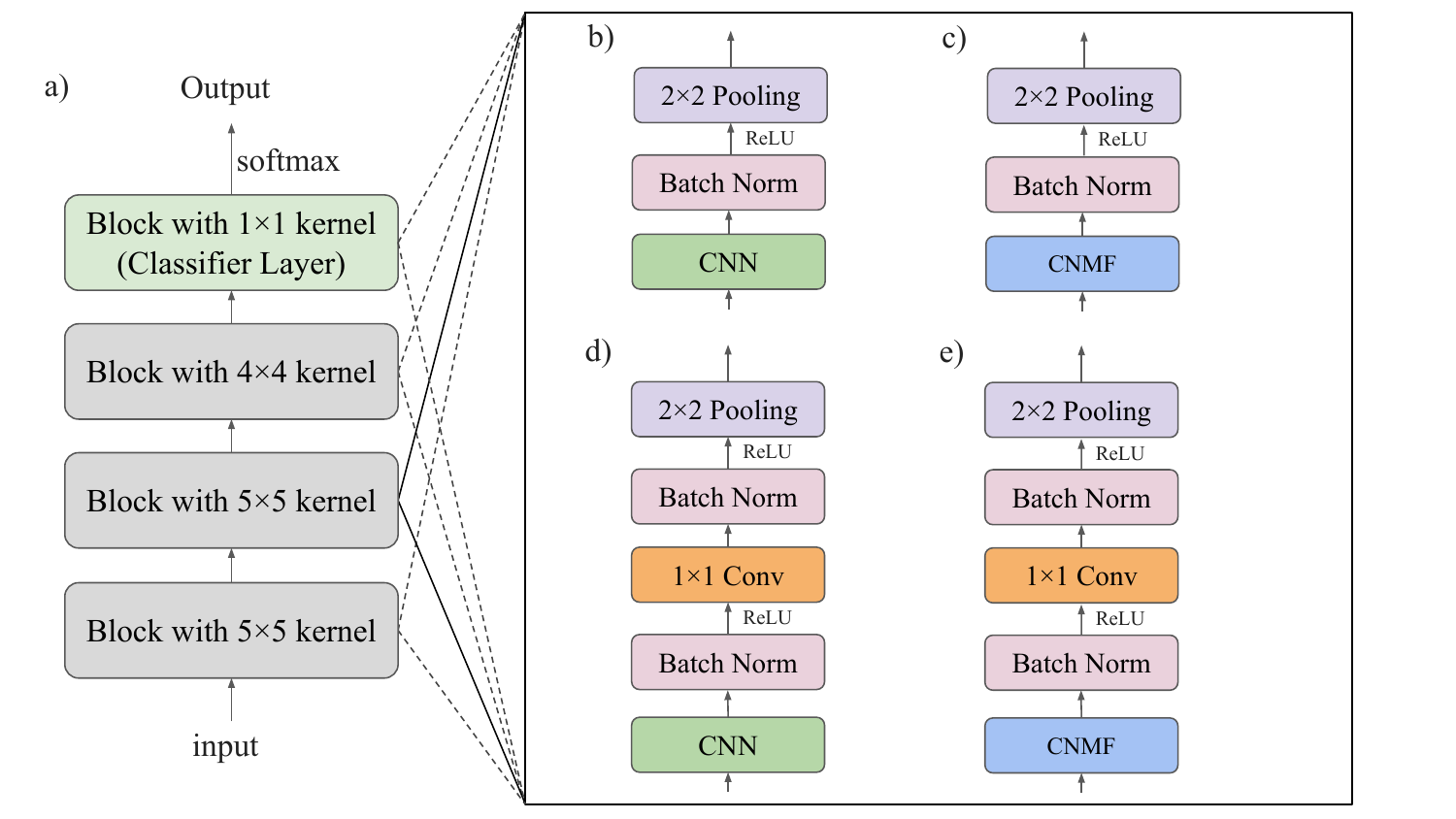}
    \caption{Model architecture of all investigated networks. a) Overall model architecture. All three convolutional layers consist of one of the modules listed on the right. b) Module used in the baseline \textbf{CNN} model. c) Module used in the \textbf{CNMF} model. d) Module used in the \textbf{CNN +  1$\times$1 Conv} model. e) Module used in the \textbf{CNMF +  1$\times$1 Conv} model.  }
    \label{cnn_models_fig}
\end{figure}

The non-negative constraint in NMF layers causes the network to represent data as a combination of basic building blocks (or "parts") that are added together, rather than canceled out. This approach excels at identifying the key components within input data. However, because NMF is fundamentally a linear method, it struggles to capture complex patterns that involve non-linear relationships between features. Our proposed architecture combines NMF convolutional layers with a layer of convolutional neural network with 1$\times$1 kernels, providing several key advantages. The subsequent 1$\times$1 convolutional layer, with its ability to use negative weights, remixes these features by allowing for subtraction and adjustment, which NMF alone cannot achieve since it can only add up contributions. This layer processes the data locally, providing detailed modulations of the more global patterns identified by the NNMF layer. 

% This combination leverages the robust feature extraction of NMF and the fine-tuning capabilities of 1$\times$1 convolutions to enhance the performance of deep convolutional networks. That is to say, the 1$\times$1 convolution essentially acts as a filter that can modulate and process the features extracted by the NMF layer with weights that can be either positive or negative.

A diagram of this module is provided in Figure \ref{cnn_models_fig} (e). We compare this model to our previous CNMF module (c) from \citep{Competitive} and its corresponding CNN model (b). We also compare this model to a similar CNN architecture shown in section (d) of the figure.

\subsection{Model Architecture}
The architecture of all proposed models consists of a sequence of four processing blocks as illustrated in Figure \ref{cnn_models_fig}a. Each block incorporates either a CNN or CNMF module, which may be followed by a $1\times 1$ convolutional layer for local feature mixing. After each layer, we apply batch normalization followed by ReLU activation.

Figure \ref{model} provides a detailed representation of our CNMF + $1\times 1 $convolution implementation (shown in Figure \ref{cnn_models_fig}e), highlighting how the architecture progressively transforms the input through successive layers. For optimization purposes, we omit batch normalization in the final two blocks.

\subsubsection{Loss Function}

To optimize our models, we employed a composite loss function that combines cross-entropy (CE) loss with mean squared error (MSE). The loss function is defined as:

\begin{equation}
    L = -\sum_i y_i \log(\hat{y}_i) + \alpha \sum_i(y_i - \hat{y}_i)^2
\end{equation}

where $y_i$ represents the true label (one-hot encoded), $\hat{y}_i$ represents the predicted probability distribution over classes, and $\alpha = 0.5$ is a weighting factor that balances the contribution of each component. While cross-entropy loss effectively optimizes for correct classification by heavily penalizing errors in the predicted class, it primarily focuses on the correct label and may not fully capture the relationship between incorrect predictions. By incorporating MSE with a smaller weight ($\alpha = 0.5$), we introduce an additional regularizing term that considers the full distribution of predictions across all classes. This combined loss function led to consistent performance improvements across all model architectures in our experiments.
\subsection{Implementation}
All models were trained using the Adam optimizer with an initial learning rate of 0.001. To ensure optimal convergence, we implemented a learning rate reduction strategy where the rate was decreased by a factor of 10 whenever the validation loss plateaued for 10 consecutive epochs. The training was terminated either when the learning rate dropped below $10^{-9}$ or when reaching the maximum limit of 500 epochs, whichever occurred first. For data augmentation, we applied random horizontal flips and color jitter to the training images. We also apply random crop on the input image from $32\times 32$ to $28\times 28$. All hyperparameters were kept consistent across different model architectures to ensure a fair comparison. The models were implemented in PyTorch and trained on NVIDIA GeForce RTX 4090 GPU.

We evaluated the performance of our proposed model on the CIFAR-10 dataset, comparing them to a CNN and NMF model similar to those described in \citep{Competitive}. The source containing all models and training setups can be found under: 
\url{https://github.com/mahbodnr/deep_nmf}

\section{Results}
\begin{figure}[htp]
    \centering
    \includegraphics[width=\textwidth]{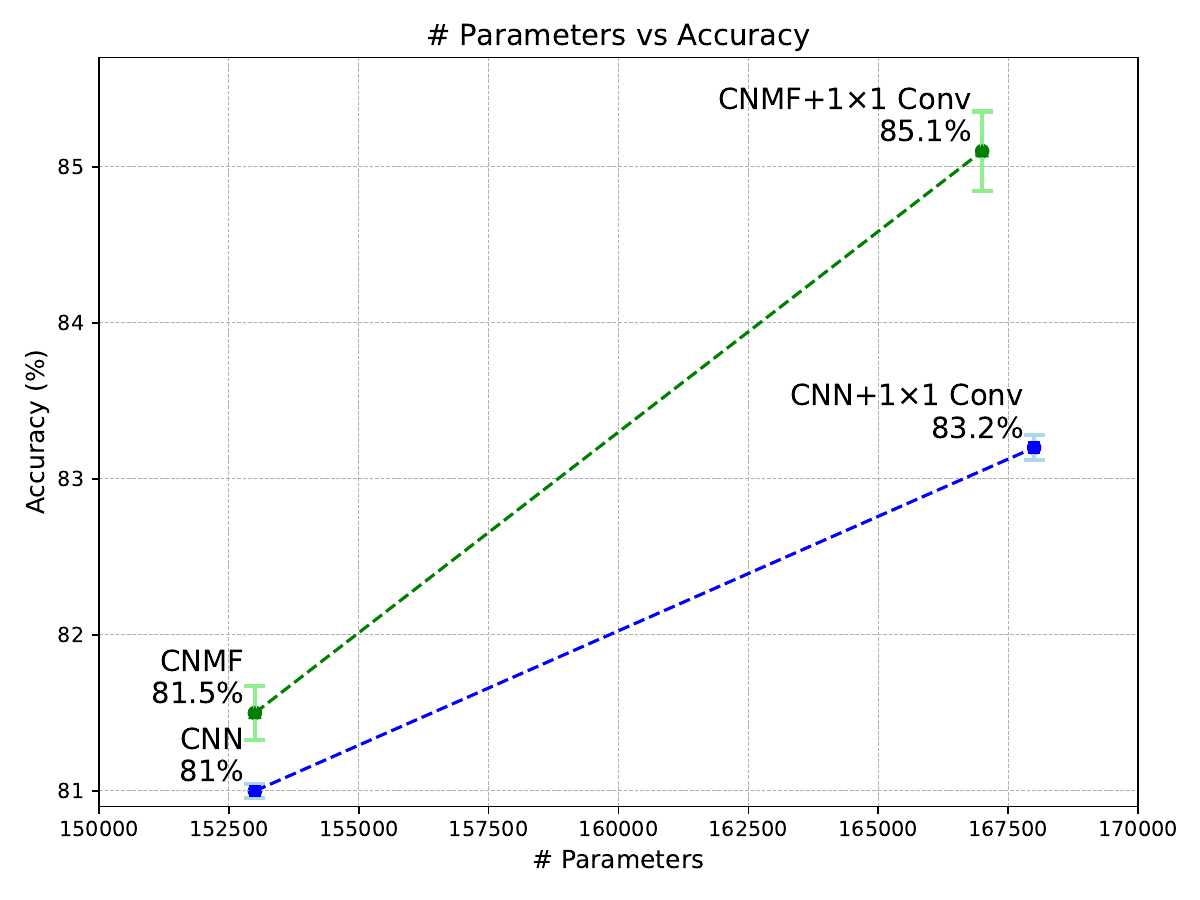}
    \caption{Classification performance of models on the CIFAR-10 dataset. Error bars represent variability across five models trained with different random initializations.}
    \label{results fig}
\end{figure}

Figure \ref{results fig} displays the classification accuracy achieved by all models on the CIFAR-10 dataset alongside their parameter counts. The results demonstrate that augmenting the CNMF model with $1\times1$ convolutions substantially improves performance, allowing it to significantly outperform the baseline CNN model of comparable architecture and size.

% \underline{link each model to figure 1}

% \begin{table}[]
% \begin{tabular}{lcc}
% \textbf{Model}                & \textbf{Number of Parameters} & \textbf{Accuracy} \\ \hline
% CNN                           & 153k                          & 81.0\%            \\
% CNN+1$\times$1 CNN            & 168k                          & 82.8\%            \\
% NMF                          & 153k                          & 81.5\%            \\
% \textbf{NMF+1$\times$1 CNN}  & 167k                          & \textbf{83.7}\%            \\
% \end{tabular}
% \label{results table}
% \caption{Models' classification performance on CIFAR-10 dataset}
% \end{table}

% \begin{figure}[htp]
%     \centering
%     \includegraphics[width=\textwidth]{accuracies.pdf}
%     \caption{Classification performance of models on the CIFAR-10 dataset. Error bars represent variability across five models trained with different random initializations.}
%     \label{results fig}
% \end{figure}

\pagebreak
\subsection{Effect of NMF compared to CNN in the network}
To investigate whether the performance improvements in our model stem primarily from the CNN components or if the NMF modules play a crucial role, we conducted an extensive analysis across different model configurations. We generated 100 different model variants by systematically adjusting the network architecture in two ways: first, by scaling the number of neurons in each layer (multiplying by factors of 2, 4, and 8), and second, by varying the number of groups in both NMF and CNN layers (using 1, 2, 4, 8, and 16 groups). When we increase the number of groups in a layer, we divide its channels into separate groups that process the input independently, thereby reducing the number of parameters while maintaining the same input and output dimensions. This approach allowed us to explore models with different ratios of NMF to CNN parameters while maintaining the overall architectural structure.

The results of this analysis are presented in Figure 3. The left panel shows model accuracy versus total parameter count, with the color intensity indicating the ratio of CNN to NMF parameters. The Pareto front, which represents the best-performing models for a given parameter budget, shows no systematic bias toward models with higher CNN parameter ratios. This suggests that simply increasing the proportion of CNN parameters does not lead to optimal performance.
The right panel provides a complementary view, plotting accuracy against the ratio of CNN to NMF parameters, with color intensity representing the total parameter count. The distribution of high-performing models appears roughly symmetric around a balanced ratio, indicating that the best results are achieved when neither component dominates the network. Notably, the highest accuracy (indicated by the red dashed line) is achieved with a nearly balanced distribution of parameters between CNN and NMF components.

These findings strongly suggest that the NMF modules are not merely passive components but are essential contributors to the network's performance. The optimal performance achieved with a balanced parameter distribution indicates a synergistic relationship between the NMF and CNN components, where each plays a crucial and complementary role in the network's processing capabilities.

% \begin{figure}[htp]
%     \centering
%     \includegraphics[width=\textwidth]{CNN_vs_NNMF_params.pdf}
%     \caption{
%     Analysis of model performance across different parameter distributions between CNN and NMF components. Left: Test accuracy versus total parameter count for 100 model variants, with color intensity indicating the ratio of CNN to NMF parameters (darker blue = higher CNN/NMF ratio). The red line shows the Pareto front of optimal-performing models. Right: Test accuracy versus CNN/NMF parameter ratio, with color intensity indicating total parameter count (darker green = more parameters). The red dashed line marks the ratio achieving highest accuracy. Models belonging to the Pareto front are indicated with a red edge. Both plots were generated by varying the number of neurons ($\times1$, $\times2$, $\times4$, $\times8$) and groups (1, 2, 4, 8, 16) in the base architecture.
%     }
%     \label{alpha_test}
% \end{figure}

\begin{figure}[htp]
    \centering
    \begin{subfigure}{\textwidth}
        \centering
        \includegraphics[width=0.9\textwidth]{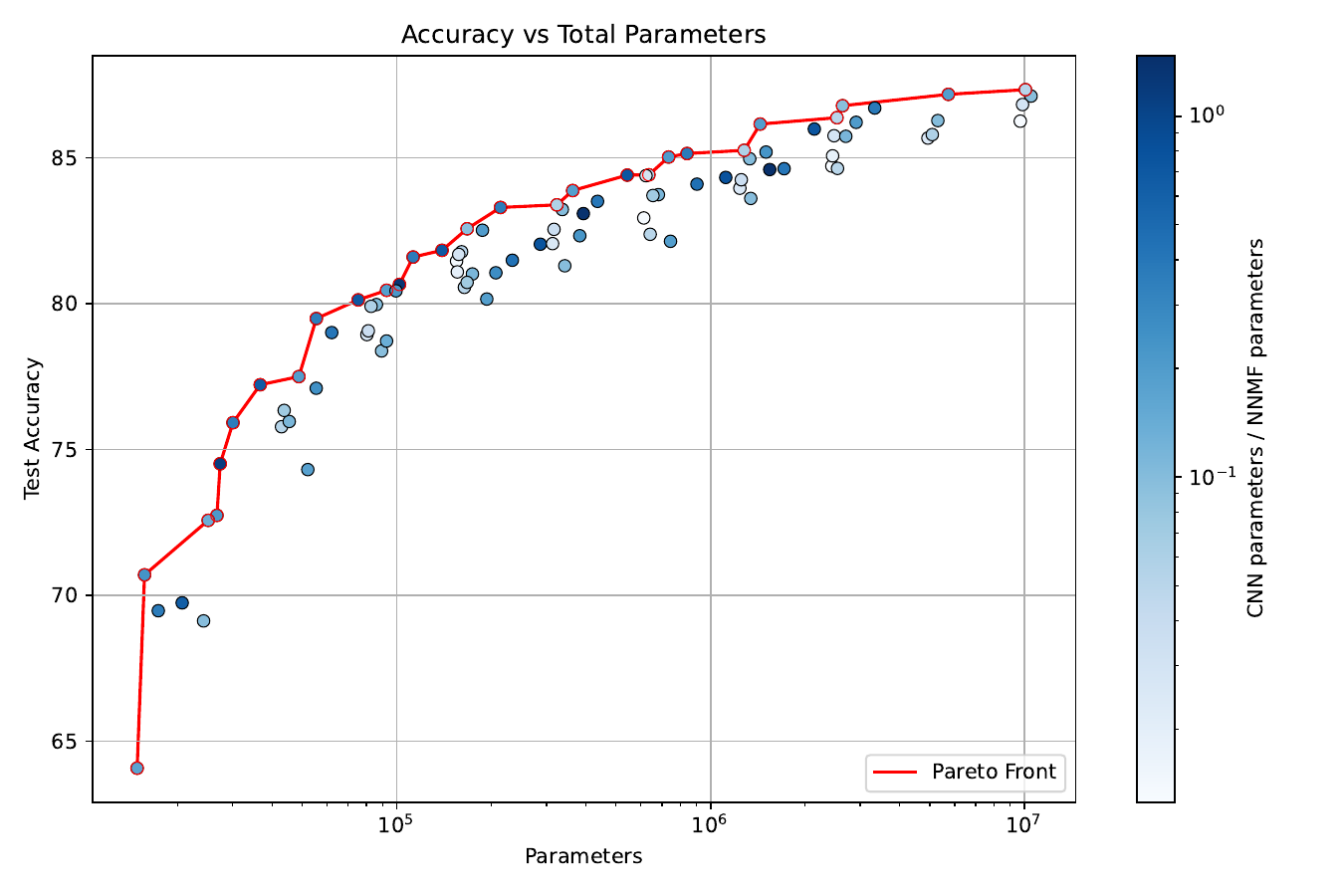}
        \caption{
        Analysis of model performance across different parameter distributions between CNN and NMF components. Left: Test accuracy versus total parameter count for 100 model variants, with color intensity indicating the ratio of CNN to NMF parameters (darker blue = higher CNN/NMF ratio). The red line shows the Pareto front of optimal-performing models.
        }
        \label{alpha_test_1}
    \end{subfigure}
    
    \begin{subfigure}{\textwidth}
        \centering
        \includegraphics[width=0.9\textwidth]{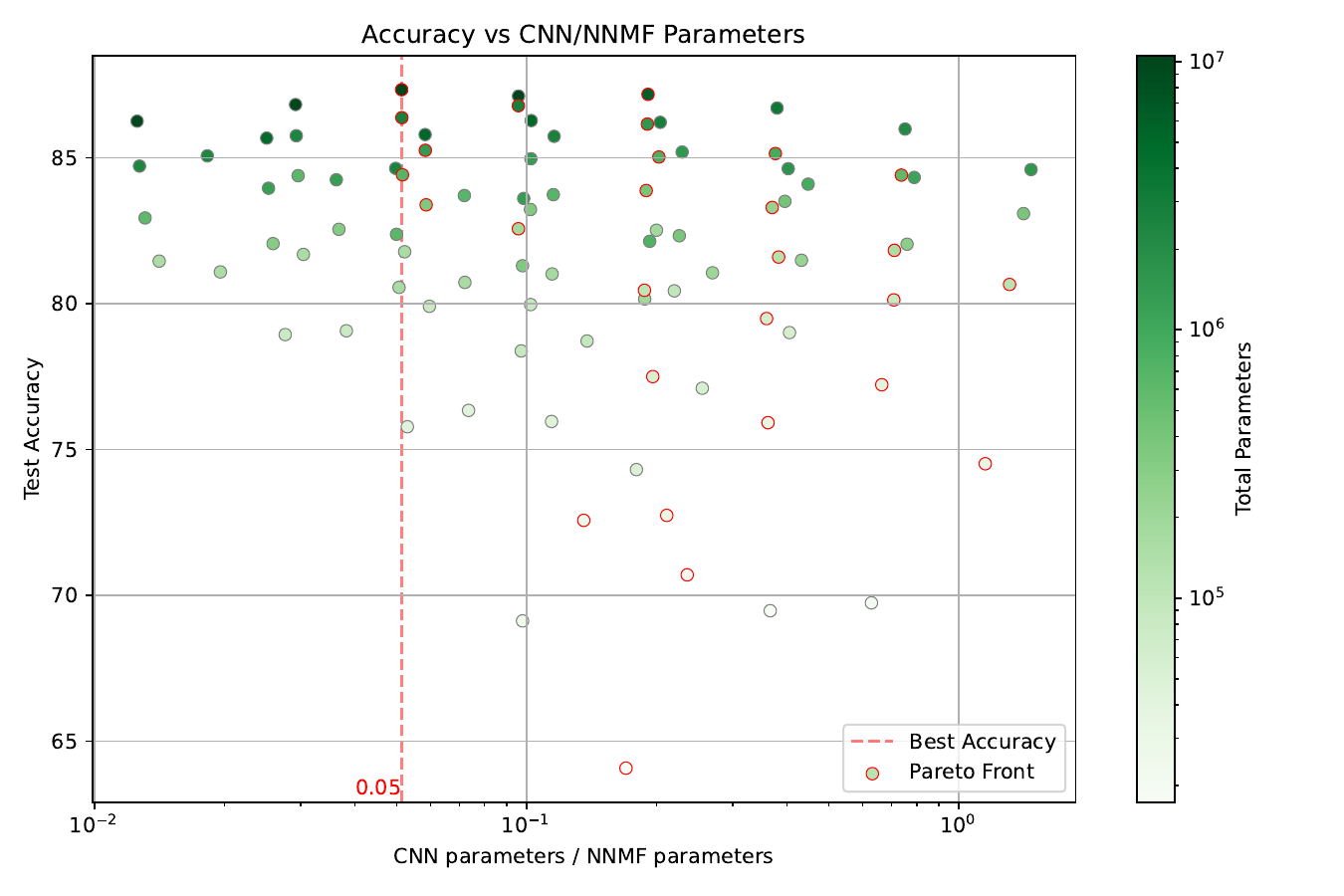}
        \caption{
         Test accuracy versus CNN/NMF parameter ratio, with color intensity indicating total parameter count (darker green = more parameters). The red dashed line marks the ratio achieving highest accuracy. Models belonging to the Pareto front are indicated with a red edge. Both plots were generated by varying the number of neurons ($\times1$, $\times2$, $\times4$, $\times8$) and groups (1, 2, 4, 8, 16) in the base architecture.
        }
        \label{alpha_test_2}
    \end{subfigure}
    
    \caption{Comparison of CNN and NMF parameter distributions.}
    \label{alpha_test}
\end{figure}

\section{Discussion and Limitations}

Our work demonstrates that incorporating biologically-inspired computational principles into deep neural networks can enhance their performance while maintaining biological plausibility. By combining NMF with local mixing through 1×1 convolutions, we achieved classification accuracy that matches or exceeds standard CNNs on the CIFAR-10 dataset, while preserving key biological constraints such as positive long-range interactions and local inhibitory processing.

\subsection{Bridging Biological and Artificial Neural Computation}

A fundamental distinction between biological neural computation and artificial neural networks lies in their computational dynamics. In biological systems, most neural processing occurs through implicit layers with complex recurrent interactions and iterative refinement of neural responses. This is evident in the cortical microcircuits where information is processed through multiple recursive loops between different neural populations before a stable representation emerges. In contrast, artificial neural networks predominantly rely on explicit feedforward computation, which, while computationally efficient, diverges significantly from biological reality.

Our approach bridges this gap by implementing NMF as an implicit layer that converges through iterative updates, more closely mimicking biological neural dynamics. While conventional feedforward networks like CNNs have dominated deep learning due to their computational efficiency and straightforward optimization, our results suggest that biologically inspired implicit computation can be equally effective when properly implemented. To make this happen, the key innovation in our work is the combination of iterative NMF processing with local feature mixing, which parallels the interaction between long-range excitatory connections and local inhibitory circuits in cortical processing.

\subsection{Analysis of Feature Selection in NMF Networks}

To understand the limitations of hierarchical NMF networks trained with unsupervised learning rules and subsequently fine-tuned for classification, it is helpful to decompose the input data into five distinct components:
\begin{itemize}
\item CD: Class-relevant Dominant statistical features
\item CN: Class-relevant Non-dominant statistical features
\item UD: class-Unrelated Dominant statistical features
\item UN: class-Unrelated Non-dominant statistical features
\item N: Noise
\end{itemize}
The key distinction between back-propagation and NMF's local learning lies in their feature selection characteristics. Back-propagation, guided by the classification objective, effectively extracts both dominant and non-dominant class-relevant features (CD and CN). In contrast, NMF's unsupervised learning rule, which optimizes for reconstruction based on statistical prominence, primarily captures dominant features regardless of their relevance to classification (CD and UD).

This fundamental difference creates a critical issue: when using NMF's local learning rules instead of back-propagation, the non-dominant but class-relevant features (CN) are progressively filtered out as information flows through the network layers. By the time the signal reaches the output layer, these crucial classification features have been lost, despite their importance for the discrimination task. This explains the reduced classification performance observed in networks trained with NMF's unsupervised learning rules. For example, a model with a similar architecture to our CNMF model will achieve $32\%$ accuracy when the NMF modules are trained only based on the local learning rule on the same task (as opposed to the $81.5\%$ that is achieved by using the back-propagation). 

This analysis highlights why our approach of using supervised gradient descent to update the weights while maintaining NMF's non-negativity constraints provides superior classification performance.

\bibliography{mybibfile}

\begin{thebibliography}{17}
\expandafter\ifx\csname natexlab\endcsname\relax\def\natexlab#1{#1}\fi
\providecommand{\url}[1]{\texttt{#1}}
\providecommand{\href}[2]{#2}
\providecommand{\path}[1]{#1}
\providecommand{\DOIprefix}{doi:}
\providecommand{\ArXivprefix}{arXiv:}
\providecommand{\URLprefix}{URL: }
\providecommand{\Pubmedprefix}{pmid:}
\providecommand{\doi}[1]{\href{http://dx.doi.org/#1}{\path{#1}}}
\providecommand{\Pubmed}[1]{\href{pmid:#1}{\path{#1}}}
\providecommand{\bibinfo}[2]{#2}
\ifx\xfnm\relax \def\xfnm[#1]{\unskip,\space#1}\fi
%Type = Article
\bibitem[{Chen et~al.(2022)Chen, Zeng and Pan}]{Chen2022}
\bibinfo{author}{Chen, W.S.}, \bibinfo{author}{Zeng, Q.}, \bibinfo{author}{Pan,
  B.}, \bibinfo{year}{2022}.
\newblock \bibinfo{title}{A survey of deep nonnegative matrix factorization}.
\newblock \bibinfo{journal}{Neurocomputing} \bibinfo{volume}{491},
  \bibinfo{pages}{305–320}.
\newblock \URLprefix \url{http://dx.doi.org/10.1016/j.neucom.2021.08.152},
  \DOIprefix\doi{10.1016/j.neucom.2021.08.152}.
%Type = Article
\bibitem[{Ciampiconi et~al.(2023)Ciampiconi, Elwood, Leonardi, Mohamed and
  Rozza}]{ciampiconi2023survey}
\bibinfo{author}{Ciampiconi, L.}, \bibinfo{author}{Elwood, A.},
  \bibinfo{author}{Leonardi, M.}, \bibinfo{author}{Mohamed, A.},
  \bibinfo{author}{Rozza, A.}, \bibinfo{year}{2023}.
\newblock \bibinfo{title}{A survey and taxonomy of loss functions in machine
  learning}.
\newblock \bibinfo{journal}{arXiv preprint arXiv:2301.05579} .
%Type = Article
\bibitem[{Fukushima(1980)}]{Fukushima1980}
\bibinfo{author}{Fukushima, K.}, \bibinfo{year}{1980}.
\newblock \bibinfo{title}{Neocognitron: A self-organizing neural network model
  for a mechanism of pattern recognition unaffected by shift in position}.
\newblock \bibinfo{journal}{Biological Cybernetics} \bibinfo{volume}{36},
  \bibinfo{pages}{193–202}.
\newblock \URLprefix \url{http://dx.doi.org/10.1007/BF00344251},
  \DOIprefix\doi{10.1007/bf00344251}.
%Type = Misc
\bibitem[{Geng et~al.(2021)Geng, Guo, Chen, Li, Wei and Lin}]{hamburger}
\bibinfo{author}{Geng, Z.}, \bibinfo{author}{Guo, M.H.}, \bibinfo{author}{Chen,
  H.}, \bibinfo{author}{Li, X.}, \bibinfo{author}{Wei, K.},
  \bibinfo{author}{Lin, Z.}, \bibinfo{year}{2021}.
\newblock \bibinfo{title}{Is attention better than matrix decomposition?}
\newblock \URLprefix \url{https://arxiv.org/abs/2109.04553},
  \DOIprefix\doi{10.48550/ARXIV.2109.04553}.
%Type = Article
\bibitem[{Hoyer(2003)}]{HOYER2003547}
\bibinfo{author}{Hoyer, P.O.}, \bibinfo{year}{2003}.
\newblock \bibinfo{title}{Modeling receptive fields with non-negative sparse
  coding}.
\newblock \bibinfo{journal}{Neurocomputing} \bibinfo{volume}{52-54},
  \bibinfo{pages}{547--552}.
\newblock \URLprefix
  \url{https://www.sciencedirect.com/science/article/pii/S0925231202007828},
  \DOIprefix\doi{https://doi.org/10.1016/S0925-2312(02)00782-8}.
  \bibinfo{note}{computational Neuroscience: Trends in Research 2003}.
%Type = Article
\bibitem[{Hubel and Wiesel(1962)}]{Hubel1962}
\bibinfo{author}{Hubel, D.H.}, \bibinfo{author}{Wiesel, T.N.},
  \bibinfo{year}{1962}.
\newblock \bibinfo{title}{Receptive fields, binocular interaction and
  functional architecture in the cat’s visual cortex}.
\newblock \bibinfo{journal}{The Journal of Physiology} \bibinfo{volume}{160},
  \bibinfo{pages}{106–154}.
\newblock \URLprefix \url{http://dx.doi.org/10.1113/jphysiol.1962.sp006837},
  \DOIprefix\doi{10.1113/jphysiol.1962.sp006837}.
%Type = Article
\bibitem[{Lecun et~al.(1998a)Lecun, Bottou, Bengio and Haffner}]{CNN}
\bibinfo{author}{Lecun, Y.}, \bibinfo{author}{Bottou, L.},
  \bibinfo{author}{Bengio, Y.}, \bibinfo{author}{Haffner, P.},
  \bibinfo{year}{1998}a.
\newblock \bibinfo{title}{Gradient-based learning applied to document
  recognition}.
\newblock \bibinfo{journal}{Proceedings of the IEEE} \bibinfo{volume}{86},
  \bibinfo{pages}{2278--2324}.
\newblock \DOIprefix\doi{10.1109/5.726791}.
%Type = Article
\bibitem[{Lecun et~al.(1998b)Lecun, Bottou, Bengio and Haffner}]{726791}
\bibinfo{author}{Lecun, Y.}, \bibinfo{author}{Bottou, L.},
  \bibinfo{author}{Bengio, Y.}, \bibinfo{author}{Haffner, P.},
  \bibinfo{year}{1998}b.
\newblock \bibinfo{title}{Gradient-based learning applied to document
  recognition}.
\newblock \bibinfo{journal}{Proceedings of the IEEE} \bibinfo{volume}{86},
  \bibinfo{pages}{2278--2324}.
\newblock \DOIprefix\doi{10.1109/5.726791}.
%Type = Article
\bibitem[{Lee and Seung(2000)}]{lee2000algorithms}
\bibinfo{author}{Lee, D.}, \bibinfo{author}{Seung, H.S.}, \bibinfo{year}{2000}.
\newblock \bibinfo{title}{Algorithms for non-negative matrix factorization}.
\newblock \bibinfo{journal}{Advances in neural information processing systems}
  \bibinfo{volume}{13}.
%Type = Article
\bibitem[{Lee and Seung(1999)}]{Lee1999}
\bibinfo{author}{Lee, D.D.}, \bibinfo{author}{Seung, H.S.},
  \bibinfo{year}{1999}.
\newblock \bibinfo{title}{Learning the parts of objects by non-negative matrix
  factorization}.
\newblock \bibinfo{journal}{Nature} \bibinfo{volume}{401},
  \bibinfo{pages}{788–791}.
\newblock \URLprefix \url{http://dx.doi.org/10.1038/44565},
  \DOIprefix\doi{10.1038/44565}.
%Type = Article
\bibitem[{Olshausen and Field(1996)}]{Olshausen1996}
\bibinfo{author}{Olshausen, B.A.}, \bibinfo{author}{Field, D.J.},
  \bibinfo{year}{1996}.
\newblock \bibinfo{title}{Emergence of simple-cell receptive field properties
  by learning a sparse code for natural images}.
\newblock \bibinfo{journal}{Nature} \bibinfo{volume}{381},
  \bibinfo{pages}{607–609}.
\newblock \URLprefix \url{http://dx.doi.org/10.1038/381607a0},
  \DOIprefix\doi{10.1038/381607a0}.
%Type = Article
\bibitem[{Olshausen and Field(2006)}]{olshausen2006other}
\bibinfo{author}{Olshausen, B.A.}, \bibinfo{author}{Field, D.J.},
  \bibinfo{year}{2006}.
\newblock \bibinfo{title}{What is the other 85 percent of v1 doing}.
\newblock \bibinfo{journal}{L. van Hemmen, \& T. Sejnowski (Eds.)}
  \bibinfo{volume}{23}, \bibinfo{pages}{182--211}.
%Type = Article
\bibitem[{Rotermund et~al.(2023)Rotermund, Garcia-Ortiz and
  Pawelzik}]{Competitive}
\bibinfo{author}{Rotermund, D.}, \bibinfo{author}{Garcia-Ortiz, A.},
  \bibinfo{author}{Pawelzik, K.R.}, \bibinfo{year}{2023}.
\newblock \bibinfo{title}{Competitive performance and superior noise robustness
  of a non-negative deep convolutional spiking network}.
\newblock \bibinfo{journal}{Neurosomething} \URLprefix
  \url{http://dx.doi.org/10.1101/2023.04.22.537923},
  \DOIprefix\doi{10.1101/2023.04.22.537923}.
%Type = Article
\bibitem[{Rotermund and Pawelzik(2019)}]{rotermund2019back}
\bibinfo{author}{Rotermund, D.}, \bibinfo{author}{Pawelzik, K.R.},
  \bibinfo{year}{2019}.
\newblock \bibinfo{title}{Back-propagation learning in deep spike-by-spike
  networks}.
\newblock \bibinfo{journal}{Frontiers in Computational Neuroscience}
  \bibinfo{volume}{13}, \bibinfo{pages}{55}.
%Type = Article
\bibitem[{Strata and Harvey(1999)}]{dale}
\bibinfo{author}{Strata, P.}, \bibinfo{author}{Harvey, R.},
  \bibinfo{year}{1999}.
\newblock \bibinfo{title}{Dale’s principle}.
\newblock \bibinfo{journal}{Brain Research Bulletin} \bibinfo{volume}{50},
  \bibinfo{pages}{349–350}.
\newblock \URLprefix \url{http://dx.doi.org/10.1016/S0361-9230(99)00100-8},
  \DOIprefix\doi{10.1016/s0361-9230(99)00100-8}.
%Type = Article
\bibitem[{Tian et~al.(2022)Tian, Su, Lauria and Liu}]{tian2022recent}
\bibinfo{author}{Tian, Y.}, \bibinfo{author}{Su, D.}, \bibinfo{author}{Lauria,
  S.}, \bibinfo{author}{Liu, X.}, \bibinfo{year}{2022}.
\newblock \bibinfo{title}{Recent advances on loss functions in deep learning
  for computer vision}.
\newblock \bibinfo{journal}{Neurocomputing} \bibinfo{volume}{497},
  \bibinfo{pages}{129--158}.
%Type = Article
\bibitem[{Yang et~al.(2013)Yang, Carrasquillo, Hooks, Nerbonne and
  Burkhalter}]{Yang2013}
\bibinfo{author}{Yang, W.}, \bibinfo{author}{Carrasquillo, Y.},
  \bibinfo{author}{Hooks, B.M.}, \bibinfo{author}{Nerbonne, J.M.},
  \bibinfo{author}{Burkhalter, A.}, \bibinfo{year}{2013}.
\newblock \bibinfo{title}{Distinct balance of excitation and inhibition in an
  interareal feedforward and feedback circuit of mouse visual cortex}.
\newblock \bibinfo{journal}{The Journal of Neuroscience} \bibinfo{volume}{33},
  \bibinfo{pages}{17373–17384}.
\newblock \URLprefix \url{http://dx.doi.org/10.1523/JNEUROSCI.2515-13.2013},
  \DOIprefix\doi{10.1523/jneurosci.2515-13.2013}.

\end{thebibliography}

\end{document}